\documentclass[prd,aps,twocolumn,eqsecnum,groupedaddress]{revtex4-2}
\usepackage[usenames,dvipsnames]{color}
\usepackage[colorlinks=true,linkcolor=Blue,citecolor=Blue,urlcolor=Blue]{hyperref}
\usepackage{graphicx}
\usepackage{bbm,bm}
\usepackage{amsmath,amsfonts,amssymb,latexsym}
\usepackage{mathptmx}
\usepackage[version=3]{mhchem}
\newcommand{\nn}{\nonumber \\}

\begin{document}
\title{DDIM Redux : Mathematical Foundation and Some Extension}
\author{Manhyung Han}
\email[Electronic address:$~~$]{hanmanhyung@gmail.com}
\affiliation{Department of Electrical and Computer Engineering, Seoul National University, Seoul 08826, Korea}

\begin{abstract} 
This note provides a critical review of the mathematical concepts underlying the generalized diffusion denoising implicit model (gDDIM) and the exponential integrator (EI) scheme. We present enhanced mathematical results, including an exact expression for the reverse trajectory in the probability flow ODE and an exact expression for the covariance matrix in the gDDIM scheme. Furthermore, we offer an improved understanding of the EI scheme's efficiency in terms of the change of variables. The noising process in DDIM is analyzed from the perspective of non-equilibrium statistical physics. Additionally, we propose a new scheme for DDIM, called the principal-axis DDIM (paDDIM).
\end{abstract}
\maketitle

\section{Introduction}
Diffusion-based generative models such as DDPM and DDIM have significantly evolved in terms of both performance and mathematical elegance over the years~\cite{prince23,dickstein15,DDPM,DDIM,song-stefano-19,song-stefano20,song21-SDE,analytic-DPM,exp-integrator22,gDDIM}. Their fundamental structures and recent extensions are well summarized in several reviews~\cite{luo22,cao22,diffusion-model-review23} and a book~\cite{prince23}. Ideas based on the score-matching concept complement the original DDPM/DDIM framework by formulating it as a stochastic differential equation (SDE) or an ordinary differential equation (ODE)\cite{song-stefano-19,song-stefano20,song21-SDE}. Recently, a more formal and mathematically rigorous understanding of diffusion models has been developed~\cite{analytic-DPM,exp-integrator22,gDDIM}.

The fundamental architecture for all diffusion probabilistic models (DPM) consists of two stages: noising and de-noising, or diffusion and inverse-diffusion. The noising (diffusion) process involves gradually scrambling the data ${\bf x}$ by adding Gaussian noise and recording the evolution of the probability $p({\bf x})$ from the initial time $t=0$ to the final time $t=T$. The probability distribution, or its score ${\bm \nabla} \log p$, is then recovered by a variational function ${\bf s}({\bf x})$, which is trained by a neural network and subsequently used to regenerate the original data in a probabilistic yet faithful manner. Score-based models allow the DPM to be formulated as an SDE and suggest ways to improve the efficiency of generation by employing the probability flow ODE~\cite{song21-SDE}. Recent works have critically analyzed how to enhance the efficiency of the generation process~\cite{exp-integrator22,gDDIM}.

This note reviews the general DPM as a diffusion process governed by two arbitrary matrices, 
${\bf f}$ and ${\bf g}$, representing the drift force and the random noise, respectively. The diffusion equation is solved exactly for a single initial point ${\bf x}(0)$, and an exact solution for the reverse trajectory is provided. The exponential-integrator scheme~\cite{exp-integrator22} and the generalized DDIM (gDDIM) scheme~\cite{gDDIM} are given a clearer mathematical exposition. The efficiency of the generation schemes noted in recent works~\cite{exp-integrator22,gDDIM} is intuitively explained. A possible generalization to the situation with multiple initial data points is attempted.

Turning to DDIM, we sharply define the condition for a diffusion scheme to qualify as DDIM and work out some exact results for the covariance matrix and other quantities. The diffusion process is then described as an approach to equilibrium, incorporating ideas from non-equilibrium statistical physics to understand the equilibration process. Finally, we propose a potentially useful scheme called the principal-axis DDIM (paDDIM) as a generalization of the well-known plain-vanilla DDIM.

In addition to aiding in improved understanding of existing DPM scheme, some original contributions made in this note are:

\begin{itemize}
    \item The diffusion and reverse-diffusion process in DPM is reviewed with added mathematical rigor for arbitrary drift (${\bf f}$) and stochastic (${\bf g}$) terms in the linear diffusion process. An exact expression for the backward trajectory is given in terms of ${\bf f}, {\bf g}$ [Eq. \eqref{general-backward-trajectory}]. 

    \item A precise condition for the diffusion process to be classified as DDIM is given as ${\bf f} = -{\bf g} {\bf g}^T /2$ [Eq. \eqref{DDIM-condition}]. Exact expression for the covariance matrix ${\bm \Sigma}(t)$ governing the diffusion process in DDIM is given [Eq. \eqref{exact-Sigma-DDIM}]. 
    
    \item The temporal evolution of the covariance matrix during the forward diffusion is analyzed from the perspective of non-equilibrium statistical physics in Sec. \ref{sec:equilibration}. 
    
\end{itemize}

Additionally, a new ansatz for the variational score function is proposed in \eqref{variational-score}; a generalization of the existing DDIM scheme involving the concept of principal axis is proposed in Sec. \ref{sec:pa-DDIM}. The validity and efficiency of these proposals will be checked in future updates. 

\section{Backgrounds}

\noindent{\bf Some Mathematical Discussion}

The backbone of diffusion models consists of two steps: forward and reverse diffusion, or noising and de-noising. The forward diffusion takes a data ${\bf x}$ in the form of a $d$-dimensional vector and adds noise to it until it becomes a random variable with zero mean. The forward process is described by SDE of the Ornstein-Uhlenbeck type, or linear SDE:
\begin{align} d{\bf x} = {\bf f}(t) {\bf x} dt + {\bf g}(t) d{\bf w} .
\label{linear-SDE}
\end{align}
The drift term ${\bf f} (t)$ and the stochastic term ${\bf g}(t)$ are both $d\times d$ dimensional, time-dependent but ${\bf x}$-independent matrices, and $d{\bf w}$ is the standard Wiener process with each component $w_i$ of the $d$-dimensional ${\bf w}$ vector following the unit normal distribution: $w_i \sim {\cal N} (0, 1)$. The same relation can be expressed in the differential equation form, 
\begin{align} \frac{d {\bf x} (t)}{dt} = {\bf f}(t) {\bf x}(t) + {\bf g}(t) {\bm \eta}(t) , 
\label{linear-SDE2} \end{align}
where $\langle \eta_i (t) \rangle = 0$ and $\langle \eta_i (t) \eta_j (t') \rangle = \delta_{ij} \delta (t-t')$. 

The initial data ${\bf x}(0)$ evolves in the data space according to 
\begin{align} {\bf x}(t) = K(t,0) {\bf x}(0) + \int_0^t dt' K(t, t') {\bf g}(t' ) {\bm \eta}(t') . 
\label{xt-solution} \end{align} 
The transition matrix $K(t,t') = U(t) U^{-1} (t')$ satisfies the composition rule $K(t_1 , t_2 ) K(t_2 , t_3 ) = K(t_1 , t_3)$. The evolution matrix $U(t)$ is the solution of 
\begin{align} \frac{d U(t)}{dt} = {\bf f}(t) U(t) \label{dUdt-main} \end{align} 
with the initial condition $U(0) = \mathbf{I}$. See Appendix \ref{app-1} for derivation of \eqref{xt-solution}.  

The expression for ${\bf x}(t)$ in \eqref{xt-solution} neatly breaks up into the mean $\langle {\bf x}(t) \rangle = K(t,0) {\bf x}(0)$ and the fluctuation $\int_0^t dt' K(t, t') {\bf g}(t' ) {\bm \eta}(t')$. From the latter part we can deduce the covariance of ${\bf x}(t)$ as 
\begin{align} 
{\bm \Sigma} (t) & = \int_0^t dt' K(t,t') {\bf D}(t') K^T (t,t')  \nn
& = U(t) \left( \int_0^t dt' U^{-1} (t') {\bf D}(t') U^{-T} (t' ) \right) U^T (t) . 
\label{Sigma-formula} \end{align} 
The diffusion matrix ${\bf D}(t)$ is defined in terms of the fluctuation matrix ${\bf g}(t)$ by 
\begin{align} {\bf D}(t) = {\bf g}(t) {\bf g}^T (t). \end{align} 
%
%In principle, one can solve for $U(t)$ by integrating \eqref{dUdt-main}, then ${\bm \Sigma}(t)$ follows from \eqref{Sigma-formula} since ${\bf D}, {\bf f}$ are known functions of time. 

The solution for ${\bf x}(t)$ in \eqref{linear-SDE2} suggests that a trajectory that began at ${\bf x}(0)$ evolves into the probability distribution for ${\bf x}(t)$ at time $t$ which is normal:
\begin{align} p ({\bf x} (t)) = {\cal N} \Bigl( {\bf x}(t); K(t,0) {\bf x} (0) , {\bm \Sigma} (t) \Bigr) . 
\label{pxt} \end{align} 
At the initial time we have ${\bm \Sigma}(0) = 0$ and the probability reduces to a delta function sharply peaked at ${\bf x} = {\bf x}(0)$. The exact score function for the forward process is given accordingly:
\begin{align} {\bm \nabla} \log p ({\bf x} (t) ) = - {\bm \Sigma}^{-1} (t) \Bigl( {\bf x}(t) - K(t,0) {\bf x} (0) \Bigr)  . \label{exact-score-main} \end{align} 
This summarizes the forward diffusion process governed by the time-dependent, matrix-valued drift term ${\bf f}(t)$ and the stochastic term ${\bf g}(t)$, for the trajectory that started at the exactly known location ${\bf x}(0)$. 
\\

\noindent{\bf Exact Solution for the Reverse Trajectory}

The corresponding reverse-diffusion process can be formulated as an SDE~\cite{anderson82,song21-SDE,diffusion-normaliziong-flow}
\begin{align} d{\bf x} & = \left( {\bf f}(t) {\bf x} - \frac{1}{2} (1+\lambda^2)  {\bf D}(t) {\bm \nabla} \log p ({\bf x}(t)) \right) dt + \lambda {\bf g}(t) d{\bf w} 
\label{general-SDE} \end{align} 
with a family of $\lambda >0$ values and $\log p({\bf x}(t))$ given by \eqref{exact-score-main}. Given the final distribution $p({\bf x}(T))$, \eqref{general-SDE} predicts the same marginal distribution $p({\bf x}(t))$ for $0 < t < T$ regardless of $\lambda >0 $. The proof, first given in \cite{diffusion-normaliziong-flow}, is reproduced in Appendix \ref{app-ODE-proof} for completeness. 

In particular, the case $\lambda =0$ is called the probability flow ODE:
\begin{align}
\frac{d{\bf x}}{dt}  = {\bf f}(t) {\bf x} + \frac{1}{2}  {\bf D}(t) {\bm \Sigma}^{-1} (t) \Bigl( {\bf x} - K(t, 0 ) {\bf x}(0 )  \Bigr)  
\label{dxdt-from-reverse-ODE-main}
\end{align}
where we have inserted the exact expression for the score given in \eqref{exact-score-main}. This equation admits a simple solution once the following change of variables is implemented:
\begin{align} {\bf x}(t) = K(t,0) {\bf x}(0) + V(t) {\bf y}(t) . 
\label{change-of-variable-from-x-to-y} \end{align}
The first term is the mean position at time $t$, and $V(t)$ is the matrix that decomposes the covariance matrix ${\bm \Sigma}(t)$ through 
\begin{align} {\bm \Sigma}(t) = V(t) V^T (t) .  \label{PSD-decomposition}  \end{align}
The two matrices ${\bm \Sigma}(t)$ and $V(t)$ are governed respectively by
\begin{align} 
\frac{d{\bm \Sigma}}{dt} & =  {\bf D} + {\bf f} {\bm \Sigma} + {\bm \Sigma} {\bf f}^T \nn 
\frac{d V}{dt} & = \frac{1}{2} {\bf D} V^{-T} + {\bf f} V . 
\label{dSigmadt-main} 
\end{align} 
See Appendix \ref{app-2} for derivation of the two equations. 

In general the choice of $V(t)$ is not unique as any orthogonal matrix $R(t)$ implementing the change $V(t) \rightarrow V(t) R(t)$ leaves the decomposition \eqref{PSD-decomposition} intact. The ambiguity is removed by treating $V(t)$ as the solution of \eqref{dSigmadt-main} with the initial condition $V(0) = 0$. Such $V(t)$ satisfying both \eqref{dSigmadt-main} and the correct initial condition is the solution of the integral equation 
\begin{align} V(t) = \frac{1}{2} \int_{0}^t  dt' K(t,t') {\bf D}(t') V^{-T} (t') . 
\label{equation-for-V}
\end{align} 
Differentiating both sides with respect to $t$ gives the differential equation for $V(t)$ in \eqref{dSigmadt-main}, and the correct boundary condition $V(0) = 0$ follows from the expression \eqref{equation-for-V}. Alternatively, \eqref{equation-for-V} is written
\begin{align} V(t) U^{-1} (t) = \frac{1}{2} \int_{0}^t  dt' U^{-1} (t') {\bf D}(t') V^{-T} (t') . 
\label{v-u-inverse}
\end{align} 
This form will become useful when we write down the formula governing the generation process in DPM. 

Remarkably, as shown in Appendix \ref{app-2}, $d{\bf y}/dt = 0$ and one can write ${\bf y} = \epsilon$ as a $d$-dimensional constant vector, admitting an exact solution 
\begin{align} {\bf x}(t) = K(t, 0 ) {\bf x}( 0 ) + V(t) \epsilon . 
\label{backward-trajectory-main}
\end{align}
This formula, while formally exact, requires a careful interpretation. It is said that the reverse-diffusion process \eqref{general-SDE} admits the same distribution $p({\bf x}(t))$ regardless of $\lambda$, including $\lambda=0$. The deterministic trajectory \eqref{backward-trajectory-main} implies the mean $\langle {\bf x}(t) \rangle = K(t,0) {\bf x}(0)$ and the covariance $V_{i\alpha}(t) V_{j\beta} (t) \langle \epsilon_\alpha \epsilon_\beta \rangle$ which, by definition, must match the covariance matrix $\Sigma_{ij} (t)$. For this relation to work, we must assume the white-noise distribution $\langle \epsilon_\alpha \epsilon_\beta \rangle = \delta_{\alpha\beta}$ for $\bf \epsilon$. By treating ${\bf \epsilon}$ as a random variable with zero mean and unit covariance, we successfully recover the correct probability distribution for ${\bf x}(t)$. This is one interpretation of the formula \eqref{backward-trajectory-main}. 

The other interpretation of it is as a deterministic path going from a fixed ${\bf x}(t)$ to ${\bf x}(0)$ in reverse time. In particular, since at large $T$ the mean position $\langle {\bf x}(T) \rangle \approx 0$, we can write
\begin{align} {\bf x}(T) \approx V(T) {\bf \epsilon} ~ \rightarrow ~ {\bf \epsilon} = V^{-1} (T) {\bf x}(T) . 
\label{fixing-e} \end{align} 
To a very good approximation we can write the solution
\begin{align} 
{\bf x}(t) = U(t) U^{-1}(0) {\bf x}( 0 ) + V(t) V^{-1} (T) {\bf x}(T) .  
\label{general-backward-trajectory}
\end{align}
We have deliberately written $K(t,0) = U(t) = U(t)U^{-1}(0)$ to emphasize the symmetry of the expressions involved. This formula provides a completely general expression for the backward trajectory connecting the initial position ${\bf x}(T)$ back to the original position ${\bf x}(0)$ in {\it any diffusion process} characterized by forward SDE \eqref{linear-SDE} and backward ODE \eqref{dxdt-from-reverse-ODE-main}, including DDIM as a special case [see Sec. \ref{sec:DDIM}]. The only assumption required for \eqref{general-backward-trajectory} to be valid is that a {\it single position} ${\bf x}(0)$ be selected to start off the diffusion process. 

%The initial position ${\bf x}(0)$ is unknown in the reverse diffusion process that kicks off with a given position ${\bf x}(T)$. It can be inferred instead from \eqref{exact-score-main}, by rewriting it
%
%\begin{align} 
%K(0,t)  {\bm \Sigma} (t)  {\bm \nabla} \log p ({\bf x} (t) ) + {\bf x}(t) = {\bf x} (0)  .
%\end{align} 

The formula \eqref{general-backward-trajectory} allows for a neat interpretation of the trajectory evolving from ${\bf x}(T)$ to ${\bf x}(0)$ as a ``straight line" connecting the two points, with relative weights given by $U(t)U^{-1}(0)$ and $V(t)V^{-1}(T)$, respectively. It is not a straight line in the literal sense of Euclidean geometry because the coefficients multiplying the two extremal positions are matrices rather than numbers. 
\\

\noindent {\bf Exponential Integrator Scheme}

Instead of writing the backward trajectory as in \eqref{general-backward-trajectory}, we can write it in a more general form
\begin{align}
{\bf x}(t) & = K(t,t_0) {\bf x} (t_0 ) + \left( \frac{1}{2} \int_{t_0}^t  dt' K(t,t') {\bf D}(t') V^{-T} (t') \right)  \epsilon   \nn 
& = K(t,t_0) {\bf x} (t_0 ) + U(t) \left( V(t) U^{-1} (t) - V(t_0 ) U^{-1} (t_0 ) \right)  {\bf \epsilon}
\label{exact-s-for-any-x0} 
\end{align}
where $0 < t_0 < T$ stands for an {\it arbitrary} initial time. (See Appendix \ref{app-3} for proof.) The second line follows from the first by utilizing the defining relation for $V(t)$ given in \eqref{v-u-inverse}. This expression is very simple as it no longer involves any integrals (though the original publication~\cite{exp-integrator22} termed it ``exponential integrator"), provided that we have prior knowledge of $U(t), V(t)$. 

For a small time interval $|t - t_0 | \ll T$, \eqref{exact-s-for-any-x0} can be used to generate the coordinate ${\bf x}(t)$ iteratively from the previous one at ${\bf x}(t_0)$. Indeed, motivated by the above exact result, \eqref{exact-s-for-any-x0}, a scheme~\cite{exp-integrator22} was proposed to generate new data over small time steps, i.e. $t=t_{s-1}, t_0 = t_s$, and $t_s - t_{s-1} \ll T$: 
\begin{align}
{\bf x}_{s-1} %& = K(t_{s-1} ,t_s ) {\bf x}_s  \nn 
%& +  \left( \frac{1}{2} \int_{t_s}^{t_{s-1}} K(t_{s-1} , \tau  ) {\bf D}(\tau ) V^{-T} (\tau ) d\tau  \right) \epsilon ({\bf x}_s ) \nn 
& = K(t_{s-1} ,t_s ) {\bf x}_s  \nn 
& +  U (t_{s-1} ) \left(  V(t_{s-1} ) U^{-1} (t_{s-1}) - V(t_s ) U^{-1} (t_s ) \right) {\bf \epsilon} ({\bf x}_s ) . 
\label{iterative-generation} 
\end{align}
Instead of constant $\epsilon$, an intricate function $\epsilon ({\bf x}_s)$ depending on the data coordinate ${\bf x}_s$ is assumed. All the quantities on the right-hand side of the equation are a priori known so that the integration procedure can proceed efficiently.  
\\

\noindent{\bf Origin of the Efficiency}

The initial state ${\bf x}(0)$ evolving according to the SDE \eqref{linear-SDE} ends up at ${\bf x}(t)$ at time $t$ with probability $p ({\bf x} (t)) = {\cal N} ( {\bf x}(t); K(t, 0) {\bf x} (0) , {\bm \Sigma} (t) )$. At time $t=0$, ${\bm \Sigma}(0) = 0$ renders a delta function for $p({\bf x}(0))$. The change of variables from ${\bf x}(t)$ to ${\bf y}(t)$ implied by \eqref{change-of-variable-from-x-to-y} amounts to rewriting this probability distribution to 
\begin{align} 
p ({\bf y} (t)) & = {\cal N} ( {\bf y}(t); 0 , \mathbf{I} ) \label{py-is-normal} \end{align} 
with zero mean and unit covariance. The change of variables has the effect of ``smoothing out" whatever irregularities and sharp fluctuations there may have been in ${\bm \Sigma}(t)$, turning it into a unit normal distribution at all times. It then makes sense to keep track of the data evolution in terms of ${\bf y}(t)$ rather than ${\bf x}(t)$, and re-express the score accordingly: 
\begin{align} 
{\bm \nabla}_{\bf y} \log p ({\bf y} ) = V^T {\bm \nabla}_{\bf x} \log p ({\bf x} ) . 
\label{score-function-for-y} 
\end{align} 
The variational function ${\bf \epsilon} ({\bf x}_s)$ in \eqref{iterative-generation} is nothing but the approximation to the ground-truth score in terms of ${\bf y}$:
\begin{align} {\bf \epsilon} ({\bf x} ) \approx - {\bm \nabla}_{\bf y} \log p({\bf y}  ) . \end{align} 
It was noted~\cite{exp-integrator22} that using ${\bf \epsilon} ({\bf x})$ rather than the bare score ${\bf s}({\bf x})$ led to improved stability in the generation process, precisely because the change of variables involved in going from ${\bf x}$ to ${\bf y}$ has the effect of removing fluctuations that might be present in ${\bm \Sigma}(t)$. 
\\

\noindent {\bf When It Fails and How to Make It Work Again}

The objective of the score function ${\bf s}_{\bm \theta} (\bf x)$ in the diffusion model is to minimize the cost function
\begin{align} & \sum_s \int d{\bf x}_s p ({\bf x}_s )  || {\bf s}_{\bm \theta} ({\bf x}_s ) - {\bm \nabla}_{{\bf x}_s} \log p ({\bf x}_s ) ||^2  = \nn
& \sum_s \int d{\bf x}_s d {\bf x}_i p ({\bf x}_s | {\bf x}_i ) p({\bf x}_i )  || {\bf s}_{\bm \theta} ({\bf x}_s ) - {\bm \nabla}_{{\bf x}_s} \log p ({\bf x}_s | {\bf x}_i ) ||^2 + C. 
\label{vincent-thm} \end{align} 
The sum $\sum_s$ is over all intermediate times $1 \le s \le N$, and the second line follows from Vincent's identity with a constant $C$ which is independent of the hyperparameter $\bm \theta$~\cite{vincent11}. 

When the initial distribution is confined to a single point, ${\bf x}_i = {\bf x}_0$, we can write $p({\bf x}_i ) = \delta ({\bf x}_i - {\bf x}_0 )$, and turn the second line in \eqref{vincent-thm} into
\begin{align} 
\sum_s \int d{\bf x}_s p ({\bf x}_s | {\bf x}_0 ) || {\bf s}_{\bm \theta} ({\bf x}_s ) - {\bm \nabla}_{{\bf x}_s} \log p ({\bf x}_s | {\bf x}_0 ) ||^2 . 
\end{align} 
The conditional probability $p({\bf x}_s | {\bf x}_0 )$ and its gradient are exactly known - see \eqref{pxt} and \eqref{exact-score-main} - and we can solve for the score function exactly without having to invoke the hyperparameters ${\bm \theta}$. 

If instead the initial distribution consists of $n$ discrete points ${\bf x}_0^{(i)}$, possibly with uneven weights $w_i$, 
\begin{align} p({\bf x}_0 ) = \frac{\sum_{i=1}^n w_i \delta ({\bf x}_i -{\bf x}_0^{(i)} ) }{\sum_i w_i } , \end{align} 
then the cost function becomes
\begin{align}
\sum_i w_i \sum_s \int d{\bf x}_s p ({\bf x}_s | {\bf x}_0^{(i)} ) || {\bf s}_{\bm \theta} ({\bf x}_s ) - {\bm \nabla}_{{\bf x}_s} \log p ({\bf x}_s | {\bf x}_0^{(i)} ) ||^2 . 
\end{align}
Though each ${\bm \nabla}_{{\bf x}_s} \log p ({\bf x}_s | {\bf x}_0^{(i)} )$ for $1 \le i \le n$ is known exactly, no single score function ${\bf s}_{\bm \theta}({\bf x}_s )$ can fit all of them simultaneously. A reasonable guess for the score function would be 
\begin{align} {\bf s}_{\bm \theta} ({\bf x}_s ) = {\bm \nabla}_{{\bf x}_s} \log  \left( \frac{ \sum_i w_i p ({\bf x}_s | {\bf x}_0^{(i)} )}{\sum_i w_i } \right) . \label{variational-score} \end{align} 
This formula reduces to ${\bf s} ({\bf x}_s ) = {\bm \nabla}_{{\bf x}_s} \log p({\bf x}_s |{\bf x}_0 )$ for a single initial point ${\bf x}_0$. In the other extreme of having performed an infinite number of sampling over ${\bf x}_0$, on the other hand, the expression inside the logarithm becomes $\int d{\bf x}_0 p({\bf x}_0 ) p({\bf x}_s | {\bf x}_0 ) = p({\bf x}_s )$, and we recover the ground truth result ${\bf s} ({\bf x}_s ) = {\bm \nabla}_{{\bf x}_s } \log p({\bf x}_s )$. The variational score function \eqref{variational-score} thus works well in the two extreme cases of a single, and an infinite number of data points, and might work well in between. 

\section{Understanding DDIM}
\label{sec:DDIM}

\noindent {\bf Definition of DDIM}

So far we have placed no constraint on the structure of ${\bf f}(t), {\bf g}(t)$ and treated them as arbitrary time-dependent vectors. The DDIM is a subset of diffusion models in which the two matrices are related by 
\begin{align}
    {\bf f}(t) = -\frac{1}{2} {\bf g}(t) {\bf g}^T (t) = -\frac{1}{2} {\bf D} (t) . 
\label{DDIM-condition} \end{align}
This includes the well-known DDIM parameterized by
\begin{align}
{\bf f} = \frac{1}{2} \frac{d \log \alpha}{dt} \mathbf{I}, ~~ {\bf g} = \sqrt{-\frac{d\log \alpha}{dt}} \mathbf{I} ,
\label{DDIM-choice} 
\end{align}
which we refer to as the {\it plain-vanilla} DDIM. The formal results we derive below are, however, valid for general DDIM defined by \eqref{DDIM-condition}. 
\\

\noindent {\bf Derivation of the Covariance Matrix}

In DDIM, the covariance matrix ${\bm \Sigma}(t)$ in \eqref{Sigma-formula} is exactly given by 
\begin{align} {\bm \Sigma }(t) = \mathbf{I} - U(t) U^T (t) 
\label{exact-Sigma-DDIM}
\end{align} 
in terms of the matrix $U(t)$ introduced in \eqref{dUdt-main}. For proof, see Appendix \ref{app-4}. At long times $U(T) \rightarrow 0$ and ${\bm \Sigma}(T) \rightarrow \mathbf{I}$. 

The positive semi-definite matrix ${\bm \Sigma}(t)$ can be decomposed as in \eqref{PSD-decomposition}, and together with \eqref{exact-Sigma-DDIM} we have
\begin{align} {\bm \Sigma} (t) = \mathbf{I} - U(t) U^T (t) = V(t) V^T (t) . \end{align}
One can find $U(t)$ as the solution of $dU/dt = {\bf f}U = -({\bf D}/2) U$ with $U(0)=\mathbf{I}$, and $V(t)$ as the solution of \eqref{equation-for-V} with $V(0) = 0$. Once $U(t)$ and $V(t)$ are solved for, \eqref{general-backward-trajectory} and \eqref{iterative-generation} can be used to generate the backward trajectory for ${\bf x}(t)$.
\\

\noindent {\bf Plain-vanilla DDIM}

The plain-vanilla DDIM given in \eqref{DDIM-choice} with the monotonic function $\alpha(t)$ satisfying $\alpha (0) = 1$ and $\alpha (T) = 0$ can be solved for $U(t)$ in \eqref{dUdt-main}: 
\begin{align} U(t) = \sqrt{\alpha(t)}, ~~ K(t,t') = \sqrt{\alpha(t)/\alpha(t')}, \end{align}
and
\begin{align} {\bm \Sigma}(t) = 1 - \alpha (t), ~~ V(t) = \sqrt{1-\alpha(t)} .
\label{U-and-V-in-DDIM} 
\end{align} 
%
%The general solution for ${\bf x} (t)$ in \eqref{exact-s-for-any-x0} becomes
%
%\begin{align}
%{\bf x}(t) 
%& = \sqrt{\frac{\alpha(t)}{\alpha(t_0)}} x(t_0) - \frac{1}{2} \int_{t_0}^t \sqrt{\frac{\alpha(t)}{\alpha(t')}} \frac{d \log \alpha(t')}{dt'} \frac{dt'}{\sqrt{1-\alpha (t')}} \epsilon  \nn 
%
%& = \sqrt{\frac{\alpha(t)}{\alpha(t_0)}} x(t_0) - \frac{\sqrt{\alpha(t)}}{2} \int \frac{d \alpha (t') }{\alpha (t') } \frac{\epsilon }{\sqrt{\alpha (t' ) \left[ 1-\alpha (t') \right] }} \nn 
%& = \sqrt{\frac{\alpha(t)}{\alpha(t_0)}} x(t_0) + \frac{\sqrt{\alpha(t)}}{2} \int \frac{d \alpha^{-1} (t')}{\sqrt{ \alpha^{-1} (t')  -1 }} \epsilon  \nn & 
%& = \sqrt{\frac{\alpha(t)}{\alpha(t_0)}} {\bf  x} (t_0) \nn 
%& + \sqrt{\alpha(t)} \left( \sqrt{ \alpha^{-1} (t)  -1 } - \sqrt{ \alpha^{-1} (t_0 )  -1 }  \right) {\bf \epsilon} 
%\label{DDIM-generation} 
%\end{align}
%after performing the requisite integration. Rewriting ${\bf x}(t) = x_{s-1}$, $\alpha(t) = \alpha_{s-1}$,  and ${\bf x}(t_0) = x_s$, $\alpha (t_0) = \alpha_s$, we recover the well-known recursion relation for ${\bf x}_s \rightarrow {\bf x}_{s-1}$ in the deterministic DDIM:
%
Well-known generative relations for ${\bf x}_s$ in DDIM are obtained by feeding $U(t), V(t)$ in \eqref{U-and-V-in-DDIM} into \eqref{iterative-generation}: 

\begin{align}
\frac{{\bf x}_{s-1}}{\sqrt{\alpha_{s-1}}} = \frac{ {\bf x}_s }{\sqrt{\alpha_s}} + \left( \sqrt{\alpha_{s-1}^{-1} -1} - \sqrt{\alpha_s^{-1} -1} \right) {\bf \epsilon} ({\bf x}_s ) . 
\label{DDIM-generation}
\end{align}

\section{Approach to equilibrium in DDIM}
\label{sec:equilibration}

The evolution of the data ${\bf x}$ through the noising process is an example of an ensemble approaching the equilibrium, and one can understood from the perspective of non-equilibrium statistical physics~\cite{KAT05,han21}. 

At equilibrium $d{\bm \Sigma}/dt =0$ and \eqref{dSigmadt-main} dictates 
${\bf f} {\bm \Sigma} + {\bm \Sigma} {\bf f}^T = -{\bf D}$. One can accordingly make the decomposition of the ${\bf f} {\bm \Sigma}$ matrix into symmetric and anti-symmetric parts: 
\begin{align}
{\bf f} {\bm \Sigma} = -\frac{1}{2} {\bf D} - {\bf Q}  ,  ~~  {\bm \Sigma}{\bf f}^T = -\frac{1}{2} {\bf D} + {\bf Q}  .
\label{sigma-at-equilibrium}
\end{align}
The anti-symmetric part ${\bf Q}^T = -{\bf Q}$ is determined by solving~\cite{KAT05}
\begin{align} {\bf f} {\bf Q} + {\bf Q} {\bf f}^T = \frac{1}{2} ({\bf f} {\bf D} - {\bf D} {\bf f}^T ) . 
\label{eq-for-Q} \end{align}
Once ${\bf Q}$ is known, one can relate the $\bf f$ matrix to the covariance matrix by
\begin{align} {\bf f} = - \left( \frac{1}{2} {\bf D} + {\bf Q} \right) {\bm \Sigma}^{-1}  . \end{align}

The probability current governing the Fokker-Planck equation $\partial_t \rho + {\bm \nabla} \cdot {\bf J} = 0$ is 
\begin{align} {\bf J} = \rho {\bf f} {\bf x} - \frac{1}{2} {\bf D} {\bm \nabla} \rho  = \rho \left( {\bf f} {\bf x} - \frac{1}{2} {\bf D} {\bm \nabla} \log p \right) . \end{align} 
Inserting ${\bf f} = - {\bf D} {\bm \Sigma}^{-1} /2$ gives ${\bf J}  =0$ provided the distribution is Boltzmann-like: $p ({\bf x}) \propto \exp ( - {\bf x}^T {\bm \Sigma}^{-1} {\bf x}/2 )$. When ${\bf Q} \neq 0$, the same equilibrium distribution holds, but now there is nonzero steady-state current
\begin{align} {\bf J}^{c} = - {\bf Q} {\bm \Sigma}^{-1} {\bf x}\rho ({\bf x}) = {\bf Q} {\bm \nabla} \rho ({\bf x}) . \end{align} 
Though this current is nonzero even at equilibrium, it is divergence-free, i.e. $\partial_i J_i^c = Q_{ik} \partial_i \partial_k \rho = 0$, and represents a {\it circulating current} that does not affect the continuity equation. From the perspective of convergence to the steady state, nonzero ${\bf J}^c$ means a particle may be trapped in a circulating pattern about the equilibrium position ${\bf x} = 0$ without being able to reach it. For efficient equilibration, it will be desirable if the circulating current was absent: ${\bf J}^c =0$. 

When ${\bf f}$ is a function of ${\bf D}$ as in DDIM, the term on the right side of \eqref{eq-for-Q} vanishes, hence ${\bf Q} = 0$ and ${\bf f} {\bm \Sigma} = -{\bf D}/2$. In particular, ${\bf f} = -{\bf D}/2$ gives ${\bm \Sigma} = \mathbf{I}$ in equilibrium in accord with the general expression \eqref{exact-Sigma-DDIM}. 
We conclude that DDIM as defined by \eqref{DDIM-condition} represents an equilibration process devoid of any circulating current - a feature that should facilitate the convergence to equilibrium. 

The decomposition \eqref{sigma-at-equilibrium} is not meaningful {\it during} the approach to equilibrium since $d{\bm \Sigma}/dt \neq 0$. Direct calculation of the probability current ${\bf J}(t)$ using $p({\bf x}(t))$ from \eqref{pxt} gives
\begin{align} {\bf J}(t) =  -\frac{1}{2} p {\bf D} \Bigl[ (\mathbf{I}-{\bm \Sigma}^{-1}(t) ) {\bf x}(t) + {\bm \Sigma}^{-1} (t) K(t,0){\bf x}(0) \Bigr] \nonumber
\end{align} 
at arbitrary time $t$. The current approach zero as $t\rightarrow T$, leaving no circulating component ${\bf J}^c$. Further analysis of the equilibration process is possible by writing the diffusion matrix ${\bf D}(t)$ and $U(t)$ in the spectral form
\begin{align} 
{\bf D}(t) & = \sum_{m=1}^d \lambda_m (t) {\bf d}_m  (t) {\bf d}^T_m (t) \nn 
U (t) & = \sum_{m, n=1}^d u_{mn}(t) {\bf d}_m (t) {\bf d}^T_n (t) 
\end{align}
using a set of time-dependent orthonormal vectors ${\bf d}_m^T (t) {\bf d}_n (t) = \delta_{mn}$ and eigenvalues $\lambda_m (t)$. In general, off-diagonal elements can be present in $U(t)$. 

Substituting both spectral representations into the evolution equation for $U$ in DDIM, i.e. $dU/dt = -({\bf D}/2) U$, gives
\begin{align}
\sum_{mn} \left(\frac{d u_{mn}}{dt} + ( {\bf u} {\bf e} )_{mn} -( {\bf e} {\bf u} )_{mn} \right) {\bf d}_m {\bf d}^T_n =  -\frac{1}{2} \sum_{mn} \lambda_m u_{mn} {\bf d}_m {\bf d}^T_n .
\label{EoM-in-d-basis} 
\end{align}
An anti-symmetric connection matrix ${\bf e}$ with elements are $e_{mn} = {\bf d}_n^T \dot{\bf d}_m = -e_{nm}$ are introduced. Matching the coefficients on both sides of \eqref{EoM-in-d-basis} gives 
\begin{align}
\frac{d u_{mn}}{dt} = -\frac{1}{2} \lambda_m u_{mn} + \sum_p ( e_{mp}u_{pn} - u_{mp}e_{pn} ) .
\label{EoM-for-uab} 
\end{align}
In the case of rigid basis, i.e. ${\bf d}_m (t) = {\bf d}_m$ and ${\bf e} = 0$, simple solutions for $u_{\alpha\beta}(t)$ are possible:
\begin{align}
u_{mn} (t) = \delta_{mn} u_m (t) , ~~  u_m (t) = \exp \left( -\frac{1}{2} \int_0^t dt' \lambda_m (t') \right)  . 
\end{align}
As a result, 
\begin{align} 
{\bm \Sigma}(t) & = 1 - U^2 (t) = \sum_m [ 1- ( u_m (t) )^2 ] {\bf d}_m {\bf d}^T_m \nn 
{\bf f}(t) {\bm \Sigma}(t) & = \sum_m \lambda_m (t) [ 1 -  ( u_m (t) )^2 ] {\bf d}_m {\bf d}_m^T \end{align}  
are both diagonal and symmetric. 
%For any ``reasonable" choice of $\lambda_\alpha (t)$ approaching a non-negative constant as $t\rightarrow T$ we can anticipate $u_m (t \rightarrow T) = 0$ and ${\bm \Sigma}(t \rightarrow T) = \mathbf{I}$ in accord with general expectations. 

It is only when we allow the rotation of the basis over time, i.e. ${\bf e}(t) \neq 0$, that anti-symmetric component ${\bf Q}(t)$ develops - see Appendix \ref{app-5} for the analysis of time-dependent basis case. In practice, ${\bf g}(t)$ and ${\bf D}(t)$ are designed to guarantee efficiency of the equilibration process and there is no reason to introduce undue complications in its structure unless a clear advantage is expected. Given that understanding, one can characterize DDIM as a diffusion-based architecture that is robust against developing an anti-symmetric component ${\bf Q}(t)$ and the circulating current~\footnote{One could choose a different basis in which the diffusion matrix ${\bf D}$ appears non-diagonal. However, such a representation is related to the diagonal one by a similarity transformation ${\bf D} \rightarrow {\bf R} {\bf D} {\bf R}^T$ where ${\bf R}$ is some orthogonal matrix, and all other matrices $U$ and ${\bm \Sigma}$ undergo the same similarity transformation. As a result, ${\bf f}{\bm \Sigma}$ transforms into ${\bf R} {\bf f} {\bm \Sigma} {\bf R}^T$, which is also symmetric.}.

\section{Principal-axis DDIM}
\label{sec:pa-DDIM}

The matrix $V(t)$, as in ${\bm \Sigma}(t) = V(t) V^T (t)$, plays a vital role in governing the reverse-diffusion trajectory, \eqref{general-backward-trajectory} and is determined by \eqref{equation-for-V}. In the spectral representation where $V(t) = \sum_m v_m (t) {\bf d}_m {\bf d}^T_m$ is also diagonal, \eqref{equation-for-V} gives
\begin{align} v_m (t) = \frac{1}{2} u_m (t) \int_{0}^t  dt' \frac{\lambda_m (t') }{u_m (t')} \frac{1}{v_m (t')} .
\label{v-m-equation} 
\end{align} 
For each component $m$ the equation is the same as in the plain-vanilla DDIM. Parameterizing $\lambda_m (t) = -d\log \alpha_m (t)/dt$ for some some set of monotone decreasing functions $\alpha_m (t)$ [$\alpha_m (0) = 1, \alpha_m (T) = 0$], we find 
\begin{align} u_m (t) = \sqrt{\alpha_m (t)} , \quad v_m (t) = \sqrt{1-\alpha_m (t) }  \end{align} 
are a self-consistent solution to \eqref{v-m-equation}. The generating relation \eqref{DDIM-generation} can also be resolved into components $x_m (t) = {\bf d}_m^T {\bf x} (t)$: 
\begin{align} 
x_m (t) 
%& = \sqrt{\frac{\alpha(t)}{\alpha(t_0)}} x(t_0) - \frac{1}{2} \int_{t_0}^t \sqrt{\frac{\alpha(t)}{\alpha(t')}} \frac{d \log \alpha(t')}{dt'} \frac{dt'}{\sqrt{1-\alpha (t')}} \epsilon  \nn 
%
%& = \sqrt{\frac{\alpha(t)}{\alpha(t_0)}} x(t_0) - \frac{\sqrt{\alpha(t)}}{2} \int \frac{d \alpha (t') }{\alpha (t') } \frac{\epsilon }{\sqrt{\alpha (t' ) \left[ 1-\alpha (t') \right] }} \nn 
%& = \sqrt{\frac{\alpha(t)}{\alpha(t_0)}} x(t_0) + \frac{\sqrt{\alpha(t)}}{2} \int \frac{d \alpha^{-1} (t')}{\sqrt{ \alpha^{-1} (t')  -1 }} \epsilon  \nn & 
& = \sqrt{\frac{\alpha(t)}{\alpha(t_0)}} x_m (t_0) \nn 
& + \epsilon_m \sqrt{\alpha_m (t)} \left( \sqrt{ \alpha_m^{-1} (t)  -1 } - \sqrt{ \alpha_m^{-1} (t_0 )  -1 }  \right)  \label{pc-DDIM} \end{align}
where $\epsilon_m  = {\bf d}_m^T {\bf \epsilon}$. 

It is conceivable that a judicious choice of the orthonormal set $\{ {\bf d}_m \}$ and the de-schedule  $\{ \alpha_m (t) \}$ could lead to faster generation of ${\bf x}_0$ from ${\bf x}_N$. Performing principal component analysis on the initial data set $\{ {\bf x}_0 \}$ to identify a small subset of relevant principal-axis vectors ${\bf d}_m$, with $m$ ranging from 1 to $d' \ll d$, could improve the efficiency of the generation process. For this reason, we refer to the scheme depicted in \eqref{pc-DDIM} as principal-axis DDIM (paDDIM). 

\section{Conclusion}
We have performed thorough mathematical re-examination of the formulations of diffusion-based models, clarified a number of mathematical details, and obtained some useful and exact formulas along the way. Numerical experiments regarding the efficiency of the proposed paDDIM scheme in Sec. \ref{sec:pa-DDIM} is under way. 

\bibliography{ref}

\appendix

\section{Proof of \eqref{xt-solution}}
\label{app-1} 
Instead of solving \eqref{linear-SDE2} directly, we write ${\bf x}(t) = U(t) {\bf y}(t)$ with $U(t)$ being 
 $d\times d$-dimensional satisfying the differential equation
\begin{align} \frac{d U(t)}{dt} = {\bf f}(t) U(t) \end{align} 
and the initial condition $U(0) = \mathbf{I}$. The solution $U(t)$ is formally expressed as the product
\begin{align} U(t) = \lim_{M\rightarrow \infty} \prod_{s=1}^{M} \exp \left( {\bf f}_s  \Delta t  \right) 
\label{U-in-product-form} \end{align} 
where ${\bf f}_s \equiv {\bf f}(t_s)$ is the matrix ${\bf f}(t)$ at the time $t=t_s$, and we have divided the time interval $[0, t]$ into $M$ steps so that $t_s = s \Delta t$ and $t=M \Delta t$. One can think of $U(t)$ as being obtained in the limit $M\rightarrow \infty$. The matrix-valued exponential $e^{ {\bf f}_s \Delta t }$ can be understood in the Trotter-Suzuki form
\begin{align} \exp ({\bf f}_s \Delta t  ) \approx 1 + {\bf f}_s \Delta t  \end{align}
since $\Delta t$ is assumed very small. It is important that in writing down the product \eqref{U-in-product-form} one must place the exponential at later time to the {\it left} of the exponential terms at earlier times. Explicitly, 
\begin{align}
U(t) = \lim_{M \rightarrow \infty} e^{{\bf f}_M \Delta t } \cdots e^{{\bf f}_{s+1} \Delta t } e^{{\bf f}_{s} \Delta t } e^{{\bf f}_{s-1} \Delta t} \cdots e^{{\bf f}_1 \Delta t} .  
\label{time-ordered-U}
\end{align}
This is known as the time-ordered product. 

In terms of ${\bf y}$ the SDE \eqref{linear-SDE2} becomes
\begin{align} \frac{d \bf y}{dt} & = U^{-1} (t) {\bf g}(t) {\bm \eta}(t) . \label{ODE-for-y} \end{align}
The inverse matrix $U^{-1}$ is constructed in a similar manner to \eqref{time-ordered-U},
\begin{align} U^{-1} (t) = \lim_{M \rightarrow \infty} e^{- {\bf f}_1 \Delta t} \cdots e^{ - {\bf f}_{s-1} \Delta t } e^{- {\bf f}_{s} \Delta t } e^{- {\bf f}_{s+1} \Delta t} \cdots  e^{- {\bf f}_M \Delta t } . 
\label{inverse-U-discrete} \end{align} 
At each finite $M$ we get $U(t) U^{-1} (t) = 1$ from taking the product of \eqref{time-ordered-U} and \eqref{inverse-U-discrete}. It also follows from the construction that $U^{-1} (t)$ satisfies the differential equation
\begin{align} \frac{dU^{-1} (t) }{dt} = - U^{-1} (t) {\bf f}(t) . \end{align} 

The drift term no longer appears in \eqref{ODE-for-y}, which can be solved immediately: 
\begin{align} 
{\bf y}(t) & = {\bf y}(0) + \int_0^t dt' U^{-1} (t') {\bf g}(t') {\bm \eta}(t') . \end{align}
Restoring ${\bf x}(t) = U(t) {\bf y}(t)$, 
\begin{align} 
{\bf x}(t) & = U(t) {\bf y}(0) + \int_0^t dt' U(t) U^{-1} (t') {\bf g}(t' ) {\bm \eta}(t')  \nn 
& = K(t,0) {\bf x}(0) + \int_0^t dt' K(t, t') {\bf g}(t' ) {\bm \eta}(t') , \end{align} 
which gives \eqref{xt-solution}. The kernel $K(t,t') \equiv U(t) U^{-1} (t')$ satisfies
\begin{align} \frac{dK (t,t') }{dt} = {\bf f} (t) K(t,t') , ~~ \frac{dK (t,t') }{dt'} = - K (t,t') {\bf f} (t') \label{dKdt} \end{align} 
and the initial condition $K (t,t)=1$. 

\section{Proof of \eqref{general-SDE}}
\label{app-ODE-proof}

We give a self-contained discussion of the Fokker-Planck (forward Kolmogorov) equation and the reverse-time Kolmogorov equation. Anderson's formula for reverse diffusion is derived, along with the generalization made in \cite{diffusion-normaliziong-flow}. For this section we write the forward diffusion process as
\begin{align} d{\bf x} = {\bf f} dt + {\bf g} d{\bf w} \label{general-SDE-app} \end{align} 
without assuming the Ornstein-Uhlenbeck form for the drift ${\bf f}$. 

The stochastic evolution of the trajectory ${\bf x}$ according to \eqref{general-SDE-app} results in the conditional probability 
\begin{align}
p({\bf x}_t | {\bf x}_{t -\Delta t } ) & = {\cal N} ({\bf x}_t ; {\bf x}_{t - \Delta t } + {\bf f}(t) \Delta t, {\bf D}(t) \Delta t )  
%\nn & = N(t) e^{ - \frac{1}{2 \Delta t} [ {\bf x}_t -  {\bf x}_{t - \Delta t } - {\bf f} \Delta t ]^T {\bf D}^{-1} [ {\bf x}_t -  {\bf x}_{t - \Delta t }  - {\bf f} \Delta t ]  } 
\nonumber \end{align} 
between a pair of trajectories $({\bf x}_{t-\Delta t }, {\bf x}_t )$ at closed spaced times $t$ and $t-\Delta t$, and ${\bf D} = {\bf g} {\bf g}^T$. The distribution $p({\bf x}_t , t)$ at time $t$ follows 
\begin{align}
p({\bf x}_t , t) = 
\int d{\bf x}_{t-\Delta t }  p({\bf x}_t | {\bf x}_{t - \Delta t} )  p({\bf x}_{t-\Delta t }, t-\Delta t)  . 
\label{C2} 
\end{align}
Explicit time dependence of the probability $p({\bf x}_t , t)$ follows from the time-varying ${\bf f}(t)$ and ${\bf g}(t)$ in the SDE. The integration over ${\bf x}_{t-\Delta t}$ will be performed after making the change of variables to ${\bf y} = {\bf x}_{t-\Delta t} + {\bf f} \Delta t -{\bf x}_t$, which results in a new integration measure $d{\bf y} = [1 + ( {\bm \nabla} \cdot {\bf f} ) \Delta t ] d{\bf x}_{t-\Delta t}$, or 
\begin{align} d{\bf x}_{t-\Delta t} = [1- ( {\bm \nabla} \cdot {\bf f}) \Delta t ] d{\bf y} . \nonumber \end{align}  
With the new variable ${\bf y}$, \eqref{C2} becomes 
\begin{align}
p({\bf x}_t , t)  & = [1- ( {\bm \nabla} \cdot {\bf f} ) \Delta t ]  \nn 
& ~~ \times \int d{\bf y}  {\cal N}({\bf y}; 0, {\bf D} \Delta t) p( {\bf x}_t + {\bf y} - {\bf f} \Delta t , t-\Delta t ) .  \nonumber 
\end{align}
The Gaussian kernel is narrowly peaked around ${\bf y} = 0$, and in the $\Delta t \rightarrow 0 $ limit one can expand the probability function inside the integral up to first order in $\Delta t$:
\begin{align}
   p({\bf x} + {\bf y} - {\bf f} \Delta t , t-\Delta t) & \approx p({\bf x} , t) - [ ( {\bf f} \cdot {\bm \nabla} ) p + \partial_t p ] \Delta t  \nn 
   & + \frac{1}{2} y_i y_j \frac{\partial^2 p }{\partial x_i \partial x_j} .
\end{align}
Terms that are linear in ${\bf y}$ and vanish after integration are not shown. Replacing $y_i y_j$ by its average $\langle y_i y_j \rangle =  D_{ij} (t) \Delta t$ gives 
\begin{align}
p = [1- ( {\bm \nabla } \cdot {\bf f} ) \Delta t ] \left[ p - ( \partial_t p  + {\bf f} \cdot {\bm \nabla} p ) \Delta t + \frac{\Delta t}{2} {\bm \nabla} \cdot {\bf D} \cdot {\bm \nabla} p \right] \nonumber 
\end{align}
where $p= p({\bf x}_t , t)$, and ${\bm \nabla} \cdot {\bf D} \cdot {\bm \nabla} = D_{ij} \partial_i \partial_j$. It is assumed that the diffusion matrix ${\bf D}$ depends on time but not on the coordinate ${\bf x}$. Collecting terms of order $\Delta t$ and setting them to zero, 
\begin{align}
& \partial_t p + ( {\bm \nabla} \cdot {\bf f} )  p + ( {\bf f} \cdot {\bm \nabla})  p - \frac{1}{2} {\bm \nabla} \cdot {\bf D} \cdot {\bm \nabla} p = 0  ~~ \Rightarrow \nn 
& \partial_t p + {\bm \nabla} \cdot \left( p {\bf f} - \frac{1}{2} {\bf D} {\bm \nabla} p \right)  = 0 . 
\label{fokker-planck} 
\end{align}
This is a continuity equation for the probability density $p$ known as the Fokker-Planck equation, with the current density ${\bf J}$ given by 
\begin{align} {\bf J} = p {\bf f} - \frac{1}{2} {\bf D} {\bm \nabla} p = p \left({\bf f} -\frac{1}{2} {\bf D} {\bm \nabla} \log p \right) . \end{align} 
Physically, the probability current consists of the drift term $p{\bf f}$ and the diffusive term $-p {\bf D} {\bm \nabla} \log p /2$. 

The probability $p({\bf x}, t)$ and the transition probability $p({\bf x}, t | {\bf x}_i , t_i )$ are related by 
\begin{align}
    p({\bf x} , t) = \int  d{\bf x}_i p({\bf x} ,t | {\bf x}_i , t_i ) p({\bf x}_i , t_i ) . 
\end{align}
The transition probability $p({\bf x} , t |{\bf x}_i , t_i )$  obeys the same Fokker-Planck equation as the probability $p({\bf x} , t)$ itself, with respect to $({\bf x}, t)$. 

The differential equation for the initial time and position $({\bf x}_i, t_i )$ of the transition probability known as the reverse Kolmogorov equation can be derived as well. One starts off with the composition rule:
\begin{align} \int d{\bf x} p ({\bf x}_f , t_f | {\bf x}, t) p({\bf x}, t | {\bf x}_i , t_i ) = p({\bf x}_f , t_f | {\bf x}_i , t_i ) . \end{align} 
Differentiating it with respect to $t$ gives 
\begin{align}
        \int d{\bf x} \left( \frac{\partial}{\partial t} p({\bf x}_f , t_f | {\bf x} , t) ) \right) p({\bf x} , t | {\bf x}_i , t_i ) \nn 
        + \int d{\bf x} p({\bf x}_f , t_f | {\bf x} , t) \left( \frac{\partial}{\partial t} p({\bf x} , t | {\bf x}_i , t_i ) \right) = 0 .
        \label{ddd}
\end{align}
We can replace $\partial_t p({\bf x}, t | {\bf x}_i, t_i )$ by $-{\bm \nabla} \cdot (p {\bf f} - {\bf D} {\bm \nabla} p /2)$ in the second integral, and perform integration by parts to rewrite \eqref{ddd} as
\begin{align} 
& \int d{\bf x} \Bigl[ \frac{\partial}{\partial t} p({\bf x}_f , t_f | {\bf x} , t) + ({\bf f} \cdot {\bm \nabla} ) p({\bf x}_f , t_f | {\bf x} , t) \nn 
& +\frac{1}{2} ( {\bm \nabla} \cdot {\bf D} \cdot {\bm \nabla} )  p({\bf x}_f , t_f | {\bf x} , t) \Bigr] p ({\bf x} , t | {\bf x}_i , t_i )   = 0 .
\end{align} 
The expression inside the bracket must vanish, yielding the backward Kolmogorov equation:
\begin{align}
\frac{\partial}{\partial t_1 } p (x_2 |x_1 ) = & -  {\bf f} (x_1) \cdot {\bm \nabla}_1 p (x_2 | x_1 ) \nn 
& - \frac{1}{2} ( {\bm \nabla}_1 \cdot {\bf D}(t_1 ) \cdot {\bm \nabla}_1 )  p (x_2 |x_1 ) ,
\end{align}
where we have expressed the space and time coordinates by $x= ({\bf x}, t)$, and $t_2 > t_1$. 

Now we are prepared to derive Anderson's reverse diffusion equation and its generalization. The probability distribution $p(x_1)$ itself follows the Fokker-Planck equation
\begin{align}
\frac{\partial}{ \partial t_1} p (x_1 ) = & -  {\bm \nabla}_1 \cdot \left(  {\bf f} (x_1) p ( x_1 ) \right)  \nn 
& + \frac{1}{2} {\bm \nabla}_1 \cdot {\bf D}(t_1 ) \cdot {\bm \nabla}_1  p ( x_1 ) . 
\end{align}
Multiplying $p(x_1)$ with $p(x_2 | x_1)$ to obtain the joint probability $p(x_1 , x_2 ) = p(x_2 | x_1 ) p(x_1 )$, we can calculate $\partial_{t_1} p(x_1 , x_2)$:
\begin{widetext}
\begin{align}
& \frac{\partial}{\partial t_1} p(x_1 , x_2 )  \nn 
= & -  p(x_1 ) {\bf f} (x_1) \cdot {\bm \nabla}_1 p (x_2 | x_1 ) - \frac{1}{2} p(x_1 ) {\bm \nabla}_1 \cdot {\bf D}(t_1) \cdot {\bm \nabla}_1 p (x_2 |x_1 ) - p(x_2 | x_1 ) {\bm \nabla}_1 \cdot \left(  {\bf f} (x_1) p ( x_1 ) \right)  +  \frac{1}{2} p (x_2 | x_1 ) {\bm \nabla}_1 \cdot {\bf D}(t_1 ) \cdot {\bm \nabla}_1 p ( x_1 ) \nn 
= & - {\bm \nabla}_1 \cdot \left( {\bf f} (x_1 ) p(x_2 | x_1 ) p(x_1 )  \right) +\frac{1}{2} \Bigl[ p (x_2 | x_1 ) {\bm \nabla}_1 \cdot {\bf D}(t_1) \cdot {\bm \nabla}_1 p ( x_1 ) - p(x_1 ) {\bm \nabla}_1 \cdot {\bf D}(t_1) \cdot {\bm \nabla}_1 p(x_2 | x_1 ) \Bigr] . 
\label{eq:179}
\end{align}
We replace $p(x_2 | x_1 ) = p(x_1 , x_2 )/p(x_1)$ and find
\begin{align} - p(x_1 ) \partial_i \partial_j \frac{p(x_1 , x_2 )}{p (x_1 ) } = & - \partial_i \partial_j p (x_1 , x_2 ) 
+ \partial_i p(x_1 , x_2 ) \partial_j \log p(x_1 ) + \partial_j p(x_1 , x_2 ) \partial_i \log p(x_1 ) - p(x_1 , x_2 ) p(x_1 ) \partial_i \partial_j  \frac{1}{p(x_1 )}  \nn 
= & - \partial_i \partial_j  p (x_1 , x_2 ) 
+ \partial_i [ p(x_1 , x_2 ) \partial_j \log p(x_1 ) ] + \partial_j [ p(x_1 , x_2 ) \partial_i \log p(x_1 ) ] - p( x_2 |x_1 ) \partial_i \partial_j  p(x_1 ) . 
\end{align}  
Inserting this into \eqref{eq:179} gives
\begin{align}
\frac{\partial}{\partial t_1} p(x_1 , x_2 ) = &  - {\bm \nabla}_1 \cdot \left( {\bf f} (x_1 ) p(x_1 , x_2 )  \right) +  {\bm \nabla}_1 \cdot [ p(x_1 , x_2 ) {\bf D} (t_1 ){\bm \nabla}_1 \log p(x_1 ) ]
 -\frac{1}{2} {\bm \nabla}_1 \cdot {\bf D} (t_1 )  \cdot {\bm \nabla}_1  p (x_1 , x_2 )  \nn 
 = & -{\bm \nabla}_1 \cdot \Bigl(p(x_1 , x_2 ) [  {\bf f} (x_1 ) - {\bf D} (t_1 ) {\bm \nabla}_1 \log p(x_1 ) ] \Bigr)  -\frac{1}{2} {\bm \nabla}_1 \cdot {\bf D} (t_1 )  \cdot {\bm \nabla}_1 p (x_1 , x_2 ) .
\end{align}
The same differential equation is satisfied by $p(x_1 |x_2 )$ and thus by $p(x_1 )$:
\begin{align}
\frac{\partial}{\partial t_1} p(x_1 ) & = -{\bm \nabla}_1 \cdot \Bigl(p(x_1 ) [  {\bf f} (x_1 ) - {\bf D} (t_1 ) {\bm \nabla}_1 \log p(x_1 ) ] \Bigr)  -\frac{1}{2} {\bm \nabla}_1 \cdot {\bf D} (t_1 )  \cdot {\bm \nabla}_1 p (x_1 )   \nn 
& = -{\bm \nabla}_1 \cdot \Bigl(p(x_1 ) [  {\bf f} (x_1 ) - \frac{1+\lambda^2}{2} {\bf D} (t_1 ) {\bm \nabla}_1 \log p(x_1 ) ] \Bigr)  -\frac{\lambda^2}{2} {\bm \nabla}_1 \cdot  \Bigl( p (x_1 ) {\bf D} (t_1 ) {\bm \nabla}_1 \log p (x_1 )  \Bigr) . 
\label{Anderson-eq} \end{align}
\end{widetext}
An arbitrary constant $\lambda^2$ is inserted for generality in the second line~\cite{diffusion-normaliziong-flow}. 

Interpretation of \eqref{Anderson-eq} should be made with care. This is still an equation with the time $t$ going {\it forward in time}. To obtain the equation for the time going backward, one must change $t\rightarrow -t$ and the two terms on the RHS of \eqref{Anderson-eq} change sign accordingly. This is the Fokker-Planck equation with the drift $- {\bf f} + \frac{1}{2}(1+\lambda^2) {\bf D} {\bm \nabla} \log p$ and the fluctuation $\lambda {\bf g}$. %The diffusive part of the current is $- \frac{1}{2} \lambda^2 {\bf D} {\bm \nabla} \log p$ and flows from the high- to low-probability region when viewed in reverse time.  
Phrased as SDE, 
\begin{align} d{\bf x} = - \Bigl[ {\bf f} - \frac{1}{2}(1+\lambda^2) {\bf D} {\bm \nabla} \log p \Bigr] dt + \lambda {\bf g} d{\bf w} , 
\end{align} 
with $dt < 0$. It is conventional to express this as 
\begin{align} d{\bf x} = \Bigl[ {\bf f} - \frac{1}{2}(1+\lambda^2) {\bf D} {\bm \nabla} \log p \Bigr] dt + \lambda {\bf g} d{\bf w} , 
\end{align} 
with $dt > 0$, which is the standard SDE form. 

\section{Proof of \eqref{backward-trajectory-main}}
\label{app-2} 

Starting from the ODE in \eqref{dxdt-from-reverse-ODE-main}
\begin{align}
\frac{d{\bf x}}{dt}  = {\bf f}(t) {\bf x} + \frac{1}{2}  {\bf D}(t) {\bm \Sigma}^{-1} (t) \Bigl( {\bf x} - K(t, 0 ) {\bf x}(0 )  \Bigr), 
\label{dxdt-original-app} 
\end{align}
we introduce a new variable ${\bf y}(t)$ by 
\begin{align} {\bf x}(t)  =  K(t, 0) {\bf x}(0) + V(t) {\bf y}(t). \label{x-to-y-app} \end{align}
where $V(t)$ decomposes the covariance matrix ${\bm \Sigma} (t) = V(t) V^T (t)$. The intuition behind this mapping is that the trajectory ${\bf x}(t)$ follows the mean $K(t,0){\bf x}(0)$ but deviates from it by an amount which is roughly the square root of the covariance, i.e. ${\bm \Sigma (t)}^{1/2} \sim V(t)$. 

The differential equation obeyed by ${\bm \Sigma}(t)$ is 
\begin{align} 
\frac{d{\bm \Sigma}}{dt} =  {\bf D} + {\bf f} {\bm \Sigma} + {\bm \Sigma} {\bf f}^T = \frac{dV}{dt} V^T + V \frac{dV^T}{dt}  . 
\label{dSigmadt} 
\end{align} 
The first equality follows from differentiating the expression for ${\bm \Sigma}(t)$ in \eqref{Sigma-formula}. The second equality follows from the definition ${\bm \Sigma}(t) = V(t) V^T (t)$. Matching the two expressions, we can deduce the differential equations for $V(t)$ by equating 
\begin{align} 
\frac{dV}{dt} V^T = \frac{1}{2} {\bf D} + {\bf f} V V^T , \quad V \frac{dV^T}{dt} =  \frac{1}{2} {\bf D} + V V^T {\bf f}^T . 
\end{align} 
Equivalently,
\begin{align} 
\frac{d V}{dt} = \frac{1}{2} {\bf D} V^{-T} + {\bf f} V , \quad \frac{dV^T}{dt} =  \frac{1}{2} V^{-1} {\bf D} + V^T {\bf f}^T .  
\label{dVdt} 
\end{align} 
This is the differential equation for $V(t)$ in \eqref{dSigmadt-main}. 

Differentiating \eqref{x-to-y-app} and equating it with \eqref{dxdt-original-app},
\begin{align} \frac{d{\bf x}}{dt} & = {\bf f}(t) {\bf x} + \frac{1}{2} {\bf D} (t) V^{-T} (t) {\bf y}(t) + V(t) \frac{d{\bf y}}{dt}.\nn 
& = {\bf f}(t) {\bf x} + \frac{1}{2} {\bf D}(t) {\bm \Sigma}^{-1} (t) V(t) {\bf y}(t) .  
\end{align} 
Since ${\bm \Sigma}^{-1} (t) V (t) = V^{-T} (t)$, we are left with the simple equation
\begin{align} V (t) \frac{d \bf y}{dt} = 0  ~ \rightarrow ~ {\bf y}(t) = \epsilon . \end{align}
This completes the derivation of the reverse-trajectory solution given in \eqref{backward-trajectory-main}:
${\bf x}(t) = K(t,0) {\bf x}(0) + V(t) \epsilon$.

\section{Proof of \eqref{exact-s-for-any-x0}}
\label{app-3}

We write the reverse-diffusion ODE \eqref{dxdt-from-reverse-ODE-main} in an equivalent form,
\begin{align}
\frac{d{\bf x}}{dt} = {\bf f}(t) {\bf x} + \frac{1}{2}  {\bf D} (t) V^{-T} (t) \epsilon ,
\label{reverse-ODE-modified-App}
\end{align}
where we have utilized the solution for ${\bf x}(t)$ in \eqref{backward-trajectory-main} to rewrite the right side of the equation. The advantage of working with this differential equation rather than \eqref{dxdt-from-reverse-ODE-main} is that it can be integrated readily from an arbitrary initial time $t_0$: 
\begin{align}
{\bf x}(t) = K(t,t_0) {\bf x} (t_0 ) + \left( \frac{1}{2} \int_{t_0}^t  dt' K(t,t') {\bf D}(t') V^{-T} (t') \right)  \epsilon  .
\end{align}
By differentiating both sides one can recover \eqref{reverse-ODE-modified-App}. Correct initial condition is recovered at $t=t_0$. 

\section{Proof of \eqref{exact-Sigma-DDIM}}
\label{app-4}
The $U(t)$ matrix satisfies the equation
\begin{align}
\frac{d U(t)}{dt} = {\bf f}(t) U(t) = - \frac{1}{2} {\bf D}(t) U(t)
\end{align}
since ${\bf F} = -{\bf D}/2$ in DDIM. This implies
\begin{align} \frac{dU^{-1}}{dt} = \frac{1}{2} U^{-1} {\bf D} , \quad \frac{dU^{-T}}{dt} = \frac{1}{2} {\bf D} U^{-T} . 
\end{align}
Then the term inside the parenthesis of \eqref{Sigma-formula} becomes
\begin{align} 
& \int_{0}^t dt' U^{-1} (t') {\bf D}(t') U^{-T} (t' ) \nn 
& = \int_{0}^t dt' \frac{d U^{-1} (t')}{dt'}  U^{-T} (t' ) + \int_{t_0}^t dt' U^{-1} (t') \frac{ dU^{-T} (t' )}{dt'} \nn 
& = \int_{0}^t dt' \frac{d}{dt'} \left[U^{-1} (t') U^{-T} (t') \right] \nn 
& = U^{-1} (t) U^{-T} (t) -1 , 
\label{integral-for-U}
\end{align} 
yielding \eqref{exact-Sigma-DDIM}.

\section{Covariance matrix in the rotating basis}
\label{app-5}

If we could ignore the rotation of the basis vectors, $d u_{\alpha\beta}/dt  = -(1/2) \lambda_\alpha u_{\alpha\beta}$ is solved readily by 
\begin{align} u^0_{\alpha\beta} = u^0_\alpha \delta_{\alpha\beta} , ~~ 
u_\alpha^0 (t) = \exp \left( -\frac{1}{2} \int_0^t dt' \lambda_\alpha (t') \right)  . 
\label{zeroth-order-u}
\end{align} 
The initial condition $U(0)=1$ is translated into $u^0_\alpha (0) = 1$ for the coefficients, since the completeness condition of the basis vectors guarantees
\begin{align} U(0) = \sum_{\alpha} {\bf d}_\alpha (0) {\bf d}^T_\alpha (0)  = 1 . \end{align} 

The zeroth-order solution is incomplete due to the rotation of the basis vectors over time, and one must write the solution in the form $u_{\alpha\beta}(t) = u_{\alpha\beta}^0 (t) + u_{\alpha\beta}^1 (t) $ with the zeroth-order solution given already in \eqref{zeroth-order-u}. The perturbative part $u_{\alpha\beta}^1 (t)$ is obtained by rewriting \eqref{EoM-for-uab} in terms of $u^0$ and $u^1$, and then isolating terms that occur at the same level of perturbation: 
\begin{align} 
\frac{du^1_{\alpha\beta}}{dt} = (u^0_\beta - u^0_\alpha) e_{\alpha\beta} . \end{align}
Its solution is 
\begin{align}
u^1_{\alpha\beta}(t) = \int_0^t dt' (u^0_\beta (t') - u^0_\alpha (t') ) e_{\alpha\beta} (t') = u^1_{\beta\alpha} (t) 
\end{align} 
and satisfies the correct initial condition $u^1_{\alpha\beta} (0) = 0$. The $U$ matrix is symmetric at this level of approximation since $u_{\alpha\beta} = u^0_{\alpha\beta} + u^1_{\alpha\beta} = u_{\beta\alpha}$. 

Using ${\bm \Sigma}(t) = 1 - U(t) U^T (t) = 1 - U^2 (t)$, matrix elements of ${\bm \Sigma}(t)$ are
\begin{align}
    \Sigma_{\alpha\beta} (t) = \left( 1- (u_\alpha^0 (t) )^2 \right) \delta_{\alpha\beta} - (u^0_\alpha (t) + u^0_\beta (t) ) u^1_{\alpha\beta} (t) , 
\end{align}
hence
\begin{align}
({\bf D} {\bm \Sigma} )_{\alpha\beta} & \approx \lambda_\alpha ( 1 - (u^0_\alpha)^2 ) \delta_{\alpha\beta} - \lambda_\alpha (u^0_\alpha + u^0_\beta ) u^1_{\alpha\beta } , \nn 
({\bm \Sigma} {\bf D}  )_{\alpha\beta} & \approx \lambda_\alpha ( 1 - (u^0_\alpha)^2 ) \delta_{\alpha\beta} - \lambda_\beta (u^0_\alpha + u^0_\beta ) u^1_{\alpha\beta } .  
\end{align}
The anti-symmetric part 
\begin{align}
    ({\bf D} {\bm \Sigma} - {\bm \Sigma} {\bf D} )_{\alpha\beta} = (\lambda_\beta - \lambda_\alpha ) (u^0_\alpha + u^0_\beta ) u^1_{\alpha\beta}
\end{align}
arises as a result of the rotation of the basis vectors. Without the rotation ${\bf D}{\bm \Sigma}$ remains symmetric at all times. 

\begin{align}
({\bm \Sigma} )_{\alpha\beta} (t) \approx \left( 1- (u_\alpha^0 (t) )^2 \right) \delta_{\alpha\beta} - (u^0_\alpha (t) + u^0_\beta (t) ) u^1_{\alpha\beta} (t) , 
\end{align}

\begin{align}
({\bm \Sigma} )_{\alpha\beta} (t) \approx \left( 1- (u_\alpha^0 (t) )^2 \right) \delta_{\alpha\beta} - (u^0_\alpha (t) + u^0_\beta (t) ) u^1_{\alpha\beta} (t) , 
\end{align}
and
\begin{align}
({\bf D} {\bm \Sigma} )_{\alpha\beta} & \approx \lambda_\alpha ( 1 - (u^0_\alpha)^2 ) \delta_{\alpha\beta} - \lambda_\alpha (u^0_\alpha + u^0_\beta ) u^1_{\alpha\beta } , \nn 
({\bm \Sigma} {\bf D}  )_{\alpha\beta} & \approx \lambda_\alpha ( 1 - (u^0_\alpha)^2 ) \delta_{\alpha\beta} - \lambda_\beta (u^0_\alpha + u^0_\beta ) u^1_{\alpha\beta }
\label{perturbative-sol} 
\end{align}
where
\begin{align}
u_\alpha^0 (t) & = \exp \left( -\frac{1}{2} \int_0^t dt' \lambda_\alpha (t') \right)  \nn 
u^1_{\alpha\beta}(t) & = \int_0^t dt' (u^0_\beta (t') - u^0_\alpha (t') ) e_{\alpha\beta} (t') = u^1_{\beta\alpha} (t) . 
\end{align}

Indeed, the anti-symmetric part is nonzero,
\begin{align}
    ({\bf D} {\bm \Sigma} - {\bm \Sigma} {\bf D} )_{\alpha\beta} = (\lambda_\beta - \lambda_\alpha ) (u^0_\alpha + u^0_\beta ) u^1_{\alpha\beta} ,
\end{align}
and arises as a result of the rotation of the basis vectors. 
\end{document}